\documentclass[twocolumn]{article}

\usepackage[utf8]{inputenc} 
\usepackage[T1]{fontenc}    
\usepackage{times}
\usepackage{hyperref}       
\usepackage{url}            
\usepackage{booktabs}       
\usepackage{amsfonts}       
\usepackage{nicefrac}       
\usepackage{microtype}      
\usepackage{fancyhdr}       
\usepackage{graphicx}       
\graphicspath{{media/}}     
\usepackage{apacite}
\usepackage{natbib}
\usepackage{amsmath}
\usepackage{listings}

\title{
\hrule\vspace{0.4cm}
Benchmarking the Rationality of AI Decision Making Using the Transitivity Axiom
\vspace{0.4cm}\hrule
}

\author{
    Kiwon Song$^{1}$, James M. Jennings III$^{1}$, Clintin P. Davis-Stober$^{1,2}$ \\
    $^1$Department of Psychological Sciences, University of Missouri \\
    $^2$University of Missouri Institute for Data Science and Informatics
}

\begin{document}
\date{}
\maketitle

\begin{abstract}
Fundamental choice axioms, such as transitivity of preference, provide testable conditions for determining whether human decision making is rational, i.e., consistent with a utility representation.  Recent work has demonstrated that AI systems trained on human data can exhibit similar reasoning biases as humans and that AI can, in turn, bias human judgments through AI recommendation systems.  We evaluate the rationality of AI responses via a series of choice experiments designed to evaluate transitivity of preference in humans. We considered ten versions of Meta's Llama 2 and 3 LLM models. We applied Bayesian model selection to evaluate whether these AI-generated choices violated two prominent models of transitivity. We found that the Llama 2 and 3 models generally satisfied transitivity, but when violations did occur, occurred only in the Chat/Instruct versions of the LLMs.  We argue that rationality axioms, such as transitivity of preference, can be useful for evaluating and benchmarking the quality of AI-generated responses and provide a foundation for understanding computational rationality in AI systems more generally.
\end{abstract}

\section{Introduction}

Recent advancements in large language models (LLMs) have driven the widespread adoption of generative artificial intelligence (AI) across various sectors to aid, or, in some cases, to replace, human decision making. Applications range from relatively simple uses, such as recommendation systems in retail, where algorithms are employed to personalize shopping experiences \citep{kunduartificial}, to more complex uses, such as analyzing credit risks, portfolio management, and fraud detection in financial management \citep{goel2023using}. AI is also used in healthcare decision making, diagnostics, and treatment planning, including the creation of novel drugs and the development of personalized patient care \citep{Jaiswaletal2020}. AI-assisted decision making is being implemented in many other domains, including agriculture, counseling, education, and government policies \citep{chen2020artificial, kuziemski2020ai, shah2022modeling, taneja2023artificial}. 

Despite the rapid integration of AI into various aspects of daily life, researchers have raised concerns about the accuracy and reliability of output generated by these systems. For instance, \cite{Athalurietal2023} identified limitations in ChatGPT’s ability to produce reliable resources for research proposals. This issue, often termed ``AI hallucination,'' refers to the information or content generated by AI that is factually incorrect, nonsensical, and/or unrelated to a given input. These errors have been observed across various platforms and can negatively influence decision-making, potentially leading to ethical and legal complications. Additionally, AI systems trained on human data can perpetuate and amplify existing societal biases. Numerous studies have shown that AI models can replicate human-like biases, including those related to gender, race, and other factors \citep{Caliskan2016SemanticsDA}. Even in critical fields like healthcare, biased AI systems have been recognized for misdiagnosing patients, compromising safety and outcomes \citep{Aquino2023MakingDB}. Biases in AI systems and AI generated false information can likewise influence human users, potentially distorting human beliefs \citep{kidd2023ai}.

To address the challenges posed by AI, numerous strategies have been developed by researchers to mitigate such AI distortions and errors. Some researchers have focused on increasing the accuracy and reliability of AI-generated outputs via knowledge graphs and checkpoints to enhance the precision of LLM models \citep{Allemang2024IncreasingTL}. Another strategy is the development of prompts designed to detect logical errors, thereby improving programming tasks such as error classification \citep{Lee2024ImprovingLC}. However, it is important to note that such errors are generally detectable only when a task is well-specified with a clear objective function and/or when responses can be clearly described as correct or not. 

We aim to evaluate the quality of the decisions or recommendations generated by AI systems directly. Building upon prior work \citep{binz2023using, hagendorff2023machine}, we adapt experimental frameworks traditionally used for studying human decision making to assess the quality and consistency of responses generated by AI systems.  Such frameworks are suitable given two key parallels between AI and human cognition. First, both can be considered, to a degree, to be ``black boxes'' with vague internal processes, where repeated observations and experiments are required to grasp a fuller understanding of the outputs \citep{barrett2020towards, norman1980twelve}. Second, both humans and many of the most widely used, and complex, AI systems generate responses in a probabilistic, i.e., non-deterministic, fashion with AI relying on probabilistic models to generate responses. 

The term \emph{rational} decision making has many meanings within various literatures.  Our work closely relates to notions of computational rationality \citep{gershman2015computational}, which describes optimal decision making by AI systems as identifying decisions with the greatest utility, subject to computational costs. We will say that a decision making process, whether human or AI-generated, is rational if it admits the existence of a utility function \cite[e.g.,][]{von1947theory} that can well-describe its responses.  Let $\mathcal{C}$ be a collection of choice alternatives (consumption set).  A \emph{utility function} $u$, is a mapping, $u: \mathcal{C}\rightarrow \mathbb{R}$, such that, for any $A, B \in \mathcal{C}$, $A$ is weakly preferred to $B$ if and only if $u(A) \geq u(B)$.  

A necessary conditions for the existence of a utility function is the \emph{transitivity of preference axiom.} Transitivity is satisfied if, and only if, for any three choice alternatives $A, B, C \in \mathcal{C}$, if $A$ is preferred to $B$ and $B$ is preferred to $C$ then $A$ is preferred to $C$.  There is broad empirical literature examining whether or not human decision makers violate transitivity in various choice contexts.  Please see \cite{ranyard2024violations,regenwetter2011transitivity,cavagnaro2014transitive,birnbaum2023testing} for recent work and empirical overviews.  Should a human or AI system violate the transitivity axiom, then there does not admit a utility representation of its responses.

Our goal is to evaluate the rationality of AI-generated responses by empirically testing whether they conform to the transitivity of preference axiom. Such a test is useful in multiple ways.  First, transitivity of preference is a normative property for decision making, e.g., the ``money pump'' argument suggests that individuals with intransitive preferences may be systematically disadvantaged or ``driven out'' of market contexts \citep{anand1993philosophy}.  Second, it allows us to better understand the consistency and structure of AI-generated responses - especially when there are not clear right/wrong answers.  Third, having information about whether or not an AI satisfies properties such as transitivity, and/or conforms to a utility model, is useful for humans who will interact with that AI.  As argued by \cite{steyvers2024three}, whether or not AI assisted human decision making is useful will depend on ``a person's collection of beliefs regarding the AI and expectations concerning the effects of interacting with the AI.''

Tests of the transitivity axiom have been previously applied to other non-human research domains, specifically that of animal behavior. For instance, \cite{arbuthnott2017mate} found that fruit flies make transitive mating choices, leading to stronger theories of fruit fly behavior.  Likewise, \cite{edwards2009rationality} examined whether ants made transitive decisions when choosing between two nest sites with varying attributes.  To be clear, we need not suppose that AI models ``reason'' as humans do. Our concern is with their generated outputs, however they are generated, and their corresponding structure. 

\section{Models of Transitivity}

While the transitivity axiom appears, at face value, to be straightforward to evaluate, there are several conceptual and technical challenges to consider.  The first to note is that the transitivity axiom is an algebraic statement, hence deterministic, and does not involve random variables, which makes it challenging to compare to systems that generate non-deterministic responses, such as humans, animals, and AI systems.  This is a well known challenge in the transitivity literature \citep{regenwetter2011transitivity,regenwetter2021ir}.  To bridge this conceptual gap, various probabilistic models of transitivity have been developed over the past 70 years, with theoretic connections to various models of choice. These models define preference in a probabilistic fashion, i.e., rather than operate at the level of preference, they consider the probability of choosing one alternative over another.

There is an obvious parallel with the LLMs we consider for the current work.  Given different randomization seeds, these models do not necessarily give identical responses upon repeated presentations of the same query. As we later demonstrate, for the LLMs we consider, even minor changes in how the choices are presented to the LLM can result in very different output. In this way, we argue that probabilistic models of transitive preference can be useful for evaluating LLM choice responses.  To clarify, we are not attempting to uncover a ``cognitive process'' underlying the LLM responses.  Rather, we can take a measurement perspective and consider support for, or against, a given probabilistic model of transitivity, as being informative of the consistency of the LLM response, i.e., what class of choice functions can well-describe the output being generated by the LLM being considered?  

For the current work, we will consider two of the most prominent probabilistic models of transitive choice: weak stochastic transitivity and the mixture model of transitive preference, which are defined as follows.

\textsc{Weak Stochastic Transitivity.} Let $P_{AB}$ be the probability of $A$ being selected over $B$, where $A$ and $B$ are members of a set, $\mathcal{C}$, of choice alternatives under consideration.  Weak stochastic transitivity (WST) holds, if and only if,
\begin{equation}\label{wst}
P_{AB} \geq \frac{1}{2} \wedge P_{BC} \geq \frac{1}{2} \Rightarrow P_{AC} \geq \frac{1}{2}, \; (\forall A,B,C \in \mathcal{C}),
\end{equation}
where ``$\wedge$'' denotes conjunction.  WST is a necessary and sufficient condition for the existence of a utility function $u$ such that
\[
u(A) \geq u(B) \Leftrightarrow P_{AB} \geq \frac{1}{2}, (\forall A,B \in \mathcal{C}).
\]
This is referred to as a weak utility model \citep{robert1979measurement}.

\textsc{Mixture model of transitive preference.}  The mixture model of transitive preference (MMTP), also referred to as the linear ordering polytope \citep{grotschel1985facets} or the random preference model of transitive choice \citep{loomes1995incorporating}, allows for an individual, whether human, animal, or LLM, to generate stochastic choice responses by randomly sampling over transitive, deterministic preferences.  Following the notion used by \cite{cavagnaro2014transitive}, let $\mathcal{T}$ be the set of all complete, asymmetric, transitive binary relations on $\mathcal{C}$.  An individual satisfies MMTP if, and only if, there exists a discrete probability distribution $\theta$ over $\mathcal{T}$ such that
\[
P_{AB} = \sum_{T \in \mathcal{T}|(A,B)\in T}\theta(T),
\]
for all $A,B \in \mathcal{C}$, where $\theta(T)$ is the probability that the individual is in transitive state $T$, and $(A,B) \in T$ denotes that $A$ is ranked ahead of (preferred to) $B$ in the relation $T$.  For choice sets $\mathcal{C}$ containing up to five distinct elements, MMTP is completely described by the following three inequalities:
\begin{equation}\label{lop1}
P_{AB}+P_{BC}-P_{AC}\leq 1 \qquad (\forall A,B,C \in \mathcal{C}),
\end{equation}
\begin{equation}\label{lop2}
P_{AB}\geq 0 \qquad (\forall A,B \in \mathcal{C}),
\end{equation}
\begin{equation}\label{lop3}
P_{AB}+P_{BA}=1 \qquad (\forall A,B \in \mathcal{C}).
\end{equation}
See \cite{regenwetter2021ir} for a recent discussion of obtaining minimal descriptions of MMTP for $|\mathcal{C}|> 5$.  It is important to note that a collection of choice probabilities, $(P_{AB})_{A,B \in \mathcal{C}, A\neq B}$, satisfying WST (resp. MMTP) does not imply that it satisfies MMTP (resp. WST).  See \cite{davis2017recasting} for a discussion and review of how WST and MMTP are necessary conditions for various classes of decision theories.

To evaluate whether LLM-generated choice responses satisfy WST and/or MMTP, we employ the order-constrained statistical methodology described in \cite{zwilling2019qtest} to calculate Bayes factors (BFs) comparing each model of transitivity (WST and MMTP) to an unconstrained benchmark model that allows for intransitive responses.  This \emph{unconstrained model} is completely described by Inequalities \eqref{lop2} and \eqref{lop3}.

A Bayes factor \citep{kass1995bayes} is the ratio of the marginal probabilities of two competing models.  For our analysis, we will calculate the following two BFs for each LLM/condition/choice set combination:
\[
BF_{WST} = \frac{\boldsymbol{D}|M_{W}}{\boldsymbol{D}|M_{U}},
\]
\[
BF_{MTP} = \frac{\boldsymbol{D}|M_{M}}{\boldsymbol{D}|M_{U}},
\]
where $\boldsymbol{D}$ is the vector of binary choice responses generated by the LLM, and $M_{W}$ and $M_{M}$ are the models formed by the binary choice probability constraints defined by Inequalities \eqref{wst} (WST) and Inequalities \eqref{lop1},\eqref{lop2},\eqref{lop3} (MMTP) respectively, with $M_{U}$ satisfying Inequalities \eqref{lop2},\eqref{lop3}. Please see \cite{zwilling2019qtest} and \cite{klugkist2007bayes} for additional details on calculating these Bayes factors.  Following \cite{jeffreys1961edition}, we consider a value of $BF_{WST}$ (resp. $BF_{MTP}$) larger than 3.16 to be ``substantial'' evidence in favor of WST (resp. MMTP), and a value of $BF_{WST}$ (resp. $BF_{MTP}$) smaller than .316 to be ``substantial'' evidence in favor of the unconstrained model, hence an intransitive choice response. If the Bayes factor is between .316 and 3.16 then the analysis is inconclusive, i.e., there is no substantial evidence to favor either model.

\section{AI models considered}
For this study, we tested 6 versions of the Llama 2 model, and 4 versions of the Llama 3 model produced by Meta. These models are provided freely by Meta for research purposes and are trained on publicly available online data. For the Llama 2 models, we used a base model (e.g., Llama-2-7b-hf) and a fine-tuned chat model (e.g., Llama-2-7b-chat-hf) with 7 billion, 13 billion, and 70 billion parameters each. Parameters in AI are adjustable variables that the model learns during training \citep{Goodfellow-et-al-2016}. Generally speaking, a larger number of parameters allows the model to learn increasingly complex patterns, potentially making more accurate predictions. Base models can have difficulty in generating tokens in expected ways without examples via techniques such as ``few-shot" or ``many-shot" prompting. The ``chat'' models build from the base models with fine-tuning from many examples with special tokens to delineate elements of a prompt, such as if a set of tokens was sent by the User or the AI during a conversation. Llama 3 models are newer and have been trained on bigger datasets and ensured to reduce false refusal rates and diversify model responses \citep{dubey2024llama}. For the Llama 3 models, we used a base and instruct (renamed from chat) model with 8 billion and 70 billion parameters each. 

Additionally, we have incorporated revision IDs for each model version to ensure transparency. This approach is critical for maintaining reproducibility, as these AI models are subject to ongoing modifications and improvements. Including the revision ID enables researchers to trace and replicate the specific version of the model used in the study. Details of the specific Llama models employed in this research are provided in Table \ref{tab:1_llama_model}. To further ensure the reproducibility of the stochastic process and facilitate unique responses by the same AI models, we randomly selected 10 numeric seeds to be used in the experiment. From these 10 seeds, we were able to obtain distinctive responses by the models. Details of the specific random seeds used in the study are provided in the Appendix. 

\begin{table*}
\centering
\begin{tabular}{lll}
\hline
\textbf{Model Version} & \textbf{Model Name} & \textbf{Revision ID} \\ \hline
Llama 2 & Llama-2-7b-hf & 01c7f73d771dfac7d292323805ebc428287df4f9 \\
Llama 2 & Llama-2-7b-chat-hf & f5db02db724555f92da89c216ac04704f23d4590 \\
Llama 2 & Llama-2-13b-hf & 5c31dfb671ce7cf2d7bb7c04375e44c55e815b1 \\
Llama 2 & Llama-2-13b-chat-hf & a2cb7a712bb6e5e736ca7f8cd89167f81a0b5bd8 \\
Llama 2 & Llama-2-70b-hf & 3aba440b59558f995867ba6e1f58f21d0336b5bb \\
Llama 2 & Llama-2-70b-chat-hf & e9149a12809580e860299585f8098ce973d1080 \\
Llama 3 & Meta-Llama-3-8B & 62bd457b6fe961a42a31306577e622c83876cb6 \\
Llama 3 & Meta-Llama-3-8B-Instruct & e1945c40cd546c78e4f1115ff4db032b271faeaa \\
Llama 3 & Meta-Llama-3-70B & b4d08b7db494d88da3ac49adf25a6b9ac01ae338 \\
Llama 3 & Meta-Llama-3-70B-Instruct & 7129260dd8548a80e1c0ace5f61c20324b472b31c\\ \hline
\end{tabular}
\caption{Llama Model and Revision ID of Each Model Used for the Experiment}
\label{tab:1_llama_model}
\end{table*}

\section{Experimental Procedure}

The experiment consisted of having each LLM complete a series of experimental trials, where each trial consisted of prompting the LLM to select a `preferred' gamble among a pair of gambles drawn from one of five stimuli sets. We applied the stimuli sets used in previous human experiments to evaluate transitivity; \cite{tversky1969intransitivity} (3 sets; \citeyear{tversky1969intransitivity}) and \cite{cavagnaro2014transitive} (2 sets; \citeyear{cavagnaro2014transitive}). Each stimuli set consisted of 5 gambles represented by a monetary value of winning and a probability of winning. All permutations of two gambles were generated for each gamble set for a total of 20 choice trials per set. With a lack of literature on observing if AI responds differently by rephrasing the same question with slight prompt changes, we constructed 6 different methods of presenting each pair of gambles. The 6 formats varied in how the probability and monetary value of winning were displayed. The probability of winning was expressed either as a fraction (e.g., 7/24) or as a percentage (e.g., 29.16\%). The monetary value of the prize was presented in three distinct ways: as a numeric value alone (e.g., 5.00), with the word ``dollar'' inserted next to the value (e.g., 5.00 dollars), or with the dollar symbol preceding the value (e.g., \$5.00). Each gamble pair/question template was presented to each of 10 LLMs and 10 randomization seed pairs for a total of 60,000 individual choice trials. Full details of the data sets and question formats are provided in Table \ref{tab:gamble_sets}.

\begin{table*}[t]
\centering
\begin{tabular}{lcc|cc|cc|cc|cc}
\hline
\textbf{Gamble Set} & \multicolumn{2}{c}{\textbf{A}} & \multicolumn{2}{c}{\textbf{B}} & \multicolumn{2}{c}{\textbf{C}} & \multicolumn{2}{c}{\textbf{D}} & \multicolumn{2}{c}{\textbf{E}} \\ 
& \textbf{Prob} & \textbf{Val} & \textbf{Prob} & \textbf{Val} & \textbf{Prob} & \textbf{Val} & \textbf{Prob} & \textbf{Val} & \textbf{Prob} & \textbf{Val} \\ \hline
Tversky 1 & 7/24 & 5.00 & 8/24 & 4.75 & 9/24 & 4.50 & 10/24 & 4.25 & 11/24 & 4.00 \\
Tversky 2 & 8/24 & 5.00 & 10/24 & 4.75 & 12/24 & 4.50 & 10/24 & 4.24 & 11/24 & 4.00 \\
Tversky 3 & 7/24 & 3.70 & 8/24 & 3.60 & 9/24 & 3.50 & 10/24 & 3.40 & 11/24 & 3.30 \\
Davis-Stober 1 & 7/24 & 25.43 & 8/24 & 24.16 & 9/24 & 22.89 & 10/24 & 21.62 & 11/24 & 20.35 \\
Davis-Stober 2 & 7/24 & 31.99 & 8/24 & 27.03 & 9/24 & 22.89 & 10/24 & 19.32 & 11/24 & 16.19 \\ \hline
\end{tabular}
\begin{minipage}{\textwidth}
\centering
\small \textit{Note.} Prob = Probability of Winning; Val = Monetary Value of Winning.
\end{minipage}

\caption{Gamble Set Used in the Experiment with Specific Probability and Monetary Value}
\label{tab:gamble_sets}
\end{table*}

We developed a Python script \footnote{All code for carrying out the experiment is available at \url{https://doi.org/10.6084/m9.figshare.28418687}} to: generate prompts from our data sets and question formats, present the prompts to the LLMs, and extract the chosen gamble from each trial.  AI models were instantiated for text-generation using the HuggingFace transformers library. Trials were performed one by one, with the randomization seed being set before each trial. This produces a memory-less environment where the choice in one trial does not influence the choice in another. For the sake of computational power, this experiment was conducted using high-performance computing infrastructure. Details of the specific computing site and resource used can be found in the Appendix. 

In our initial findings, some base models would hallucinate and produce responses unfit for the study. Therefore, to restrict the models from hallucinating, we constrained the models to respond with only a single token corresponding to the model’s preference between the two gambling pairs. Example of the hallucination can be found in the appendix. Restricting the response to only one token may raise problems due to restriction of reasoning. It has been shown that if an AI system generates reasoning before generating the answer, it can yield more accurate answers to problems with a correct solution \citep{xie2024calibrating}.

From the 60,000 responses collected in our trials, the data were initially divided by question format, creating six distinct datasets. Responses were then aggregated by LLM and gamble data set across the 10 random seeds. This process resulted in a total of 20 vectors of choice sets for each Llama model, where each cell could yield 10 possible preferences between a given choice pair. For example, the vector AB referred to the count of instances where Choice A was preferred over Choice B. Within the dataset, permuted choice sets were further aggregated across questions that included identical choices presented in different orders. For instance, the vectors AB and BA posed identical questions but in reversed order. These were aggregated into a single vector, AB, representing the preference for Choice A over Choice B. This aggregation process produced a total of 10 columns of choice sets for each row, corresponding to each Llama model and gamble set. For instance, for a question format where probabilities of winning were depicted as fractions, and monetary value as a numeric value alone, the vector AB = [20, 10, 16, 12 ... ] where each value represented each Llama model's response to a given gamble set. Using this aggregated data, Bayes factors were calculated for each Llama model based on the binary choice responses it generated for each stimuli set and presentation type combination.

\section{Results}

In all, a total of 600 Bayes factors were calculated: $BF_{WST}$ and $BF_{MMTP}$ were each calculated for all combinations of 10 LLM models, 5 stimuli sets, across 6 presentation types.  We now summarize these results to answer the following four questions:
\begin{enumerate}
    \item Which model of transitivity, WST or MMTP, was most often violated?
    \item Which AI model was most frequently associated with intransitive responses?
    \item Which question format and stimuli set led to the highest rate of intransitive responses?
    \item Which model of transitivity performed the best overall?
\end{enumerate}

\begin{figure*}[h]
    \centering
    \includegraphics[width=0.8\textwidth]{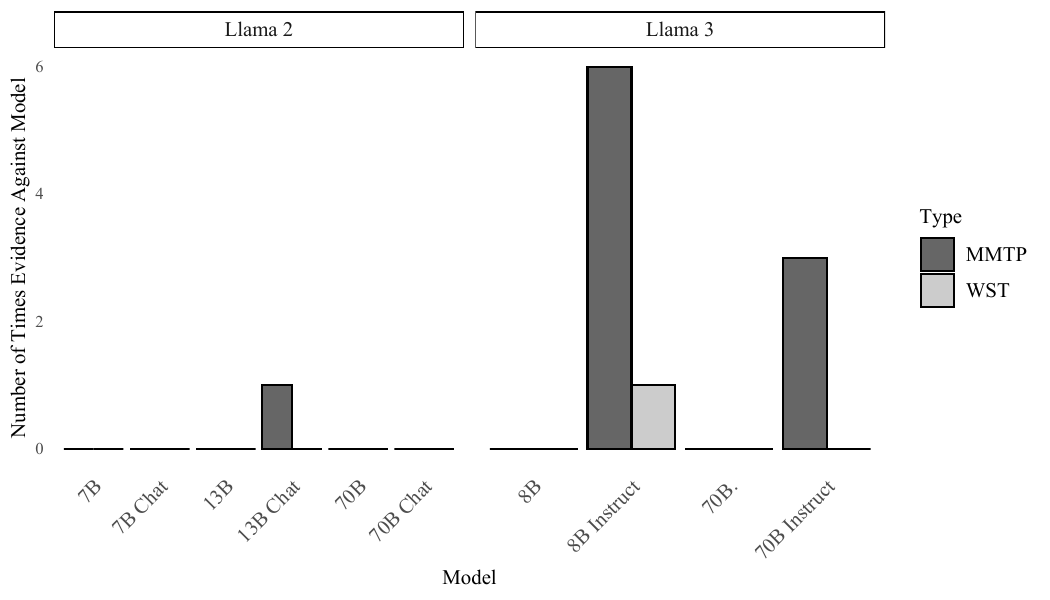}
    \vspace{1ex}
    \begin{minipage}{0.8\textwidth} 
        \small \textit{Note.} MMTP = Mixture Model of Transitive Preferences, WST = Weak Stochastic Transitivity. The x-axis refers to the parameter sizes for each Llama Model. A failure of model is counted when BF is less than .316.  
    \end{minipage}
    \caption{Substantial Evidence Against Transitivity Models by Llama Versions}
    \label{fig:1_Intransitive}
\end{figure*}

To address the first two questions, Figure \ref{fig:1_Intransitive} reports the number of transitivity violation instances observed across the gamble sets. Substantial evidence favoring the unconstrained model is indicated when the Bayes factor falls below the threshold of 0.316. We define a ``violation'' for the transitive model when this criterion is met. The figure displays the counts of violation instances for each Llama model, evaluated across 5 gamble sets and 6 question formats, resulting in 30 opportunities for failure per model. Overall, the results indicate that substantial evidence favoring the unconstrained model is rare, with only 11 failure instances observed across all gamble sets. Notably, the MMTP model demonstrated a higher number of failures compared to the WST model. Specifically, MMTP reported 10 failure instances, while WST recorded only one. This discrepancy may be due to MMTP’s stricter constraints, which could account for its higher rate of failure instances.

Furthermore, the results show that a violation of a transitivity model occurred in 7 out of 11 instances for the Llama 3 8B Instruct model. Interestingly, only the chat and instruct versions of the models exhibited intransitivity. It is interesting that the fine-tuned models failed to be transitive in these instances, while the base models did not. 

To address the third question, Table \ref{tab:3_Qustion_Format} reports the frequency of transitivity failures by question format. The rows correspond to the five gamble sets, while the columns represent the six formatting conditions used to display the probabilities and monetary values of winning. Probabilities were formatted either as fractions or percentage, and monetary value were displayed as numeric value alone (None) with the word ``dollars'' appended (Dollars), or with the dollar sign preceding the value (\$). Each cell in the table reports the number of failures observed for the corresponding combination of probability format and monetary value presentation within each gamble set. With two transitivity models analyzed across 10 different Llama models, each cell can yield up 20 violations. The results reveal that the format of the question—despite maintaining consistent contextual information—can potentially influence AI outputs. While there are only a few instances, the data show that transitivity models most often failed when probabilities of winning were presented in percentage formats. Specifically, 6 of the 11 observed failures occurred when monetary values were denoted with a dollar sign \$ in front of the value and the probability of winning was presented as a percentage.

\begin{table}[h]
    \centering
    \small  
    \begin{tabular}{|l|ccc|ccc|}
        \hline
        \textbf{Experiments} & \multicolumn{3}{|c|}{\textbf{Fraction}} & \multicolumn{3}{c|}{\textbf{Percentage}} \\
        & \textbf{None} & \textbf{Dollars} & \textbf{\$} & \textbf{None} & \textbf{Dollars} & \textbf{\$} \\
        \hline
        Gamble Set 1 & 1 & 1 & 0 & 0 & 0 & 0 \\
        Gamble Set 2 & 0 & 0 & 0 & 1 & 0 & 3 \\
        Gamble Set 3 & 0 & 0 & 0 & 0 & 0 & 1 \\
        Gamble Set 4 & 0 & 1 & 0 & 0 & 1 & 1 \\
        Gamble Set 5 & 0 & 0 & 0 & 0 & 0 & 1 \\
        \hline
        \textbf{Total} & \textbf{1} & \textbf{2} & \textbf{0} & \textbf{2} & \textbf{1} & \textbf{6} \\
        \hline
    \end{tabular}
    \begin{minipage}{\linewidth}
        \small
        \textit{Note.} Experiments: Gamble Set 1 = Davis-Stober set 1, Gamble Set 2 = Davis-Stober set 2, Gamble Set 3 = Tversky set 1, Gamble Set 4 = Tversky Set 2, Gamble Set 5 = Tversky Set 3.
    \end{minipage}
    
    \caption{Number of Times There Was Substantial Evidence Against Model of Transitivity By Question Format}
    \label{tab:3_Qustion_Format}
\end{table}

\begin{figure*}[ht]
    \centering
    \includegraphics[width=0.8\textwidth]{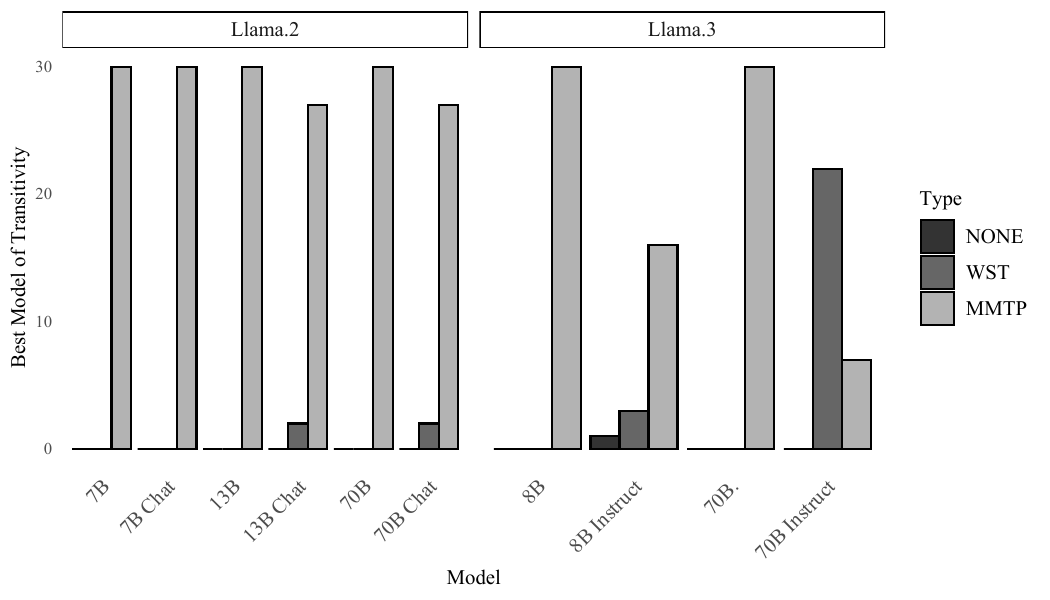}
    \vspace{1ex}
    \begin{minipage}{0.8\textwidth} 
        \small 
        \textit{Note.} NONE = Neither of the Models of Transitivity was best, MMTP = Mixture Model of Transitive Preferences, WST = Weak Stochastic Transitivity. The x-axis refers to the parameter sizes for each Llama Model.
    \end{minipage}
    \caption{Number of Times Each Transitivity Model was Best According to Bayes Factor by Llama Versions}
    \label{fig:2_Best}
\end{figure*}

The last question is addressed using the results presented in Figure \ref{fig:2_Best}, which displays the number of times each transitivity model identified as the best based on Bayes factor analysis. Among all tested Llama models, MMTP was identified as the preferred transitivity model in the majority of cases. However, WST was favored over MMTP in one instance, specifically with the Llama 3 70B Instruct model. Additionally, simultaneous violations of both transitivity models occurred in only one Llama model: the Llama 3 8B Instruct model. This occurred specifically when the question format combined probabilities expressed as percentages with monetary values denoted by a dollar sign (\$). This format, identified in Table \ref{tab:3_Qustion_Format}, produced the highest number of failures. 

The figure also highlights notable variations in transitivity model selection based on different versions of AI model. Base models demonstrated a consistent preference for MMTP, while Chat and Instruct models displayed greater variability in selecting the best transitive model. Interestingly, these variations were particularly prominent in larger-parameter Llama models, and were more pronounced in Llama 3 models compared to Llama 2 models.

Across findings from the Bayes factor analysis, the results underscore the key patterns of transitivity models across different AI models and conditions. MMTP is shown as the best transitive model under most scenarios, despite its stricter constraints compared to WST. Failure to meet the constraints of transitive models varied across AI models, however, the analysis showed that Llama 3 models, and specifically fine-tuned instruct models were accountable for most of the failures and variations in selecting the best transitive model. 

In this study, we cannot conclusively determine the underlying explanation for the violations of the transitivity models and variation observed in the responses generated by the Llama models. However, we hypothesize that the fine-tuning process of the Llama models may have contributed to the observed patterns. The Chat and Instruct models have been reported to be fine-tuned for specific purposes. The Chat models are designed to be more ``interactive,'' and prioritize on helpfulness and safety of their response before informativeness \citep{touvron2023llama}. Similarly, Instruct models are prioritized to follow directive quires and formal instructions by the user \citep{dubey2024llama}. These modifications may have prioritized certain functionalities over others, potentially leading to the observed differences and failure of the transitive models. Furthermore, both fine-tuned models inherently adjust their response patterns to be more varied and contextually appropriate to user input. As a result, compared to the base models, fine-tuned models may have produced more variable outputs when responding to the same question. This variability could account for the observed variations in selecting the best transitivity model.

\section{Discussion}

We examined the rationality of AI-generated choice through the use of an experimental choice framework developed for evaluating human choice.  The LLMs we investigated largely satisfied transitivity, with all but one model type (Llamma-3 70 billion parameters, Instruct) generally favoring a mixture-model of transitive preference.  Several LLM/choice set combinations did lead to strong violations of transitivity, with all such ``non-rational'' responses generated from Chat/Instruct LLMs.

For the current work, we chose to focus on choice sets comprised of simple gambles that have been extensively applied to examining transitivity in human participants \cite[e.g.,][]{tversky1969intransitivity}. In future work, it would be valuable to investigate how other rationality properties - such as regularity, random utility \cite[e.g.,][]{mccausland2020testing}, or other rational reasoning frameworks - hold or fail in AI systems. Such investigation may reveal parallels to human decision-making biases, e.g., \cite{li2025actionsspeaklouderwords} found AI models to exhibit implicit biases in outputs reflecting sociodemographic biases.  Going further, one could explore the degree to which stochastic variability in AI responses is related to confidence and/or strength of preference, as has been examined in human decision making \citep{alos2021choice}.  One could also consider the degree to which classic context effects impact AI decision making (e.g., similarity effect, asymmetric dominance - see \cite{davis2023illustrated} for a recent overview).

Future work could adapt our framework to more general choice settings, such as those where AI decision systems are currently being deployed/evaluated. Recent work in medical decision making has highlighted the need for LLMs to incorporate metacognition principles into evaluative frameworks for improving the quality and explainability of the resulting AI decision systems \citep{griot2025large}. Traditional principles of rational decision making, such as transitivity, could play a role in such evaluative frameworks.


Future work could also explore testing rational decision making in human-AI hybrid decision making systems \citep{schoenegger2024wisdom,steyvers2022bayesian}. Such work could determine whether collaborative approach mitigate deviations from rationality principles, and under which circumstances humans utilize the usage of AI in their decision process. 

While the current work used a memory-less environment, this framework could be applied to choice preferences across a history of choices making use of the large context sizes found in newer large language models. This perspective may reveal whether AI preferences remain consistent or whether it shifts preferences as it processes more information. A related question is whether rationality constraints - such as transitivity - should be enforced in AI systems. Although doing so may enhance the perceived reliability and consistency of recommendations, it might also impose limitations that are unhelpful in certain contexts. It is worth noting that the models of transitivity we considered place strong constraints on the space of allowable choice probabilities\footnote{For five choice alternatives, only 12\% and 5\% of all possible binary choice probabilities satisfy WST and MMTP respectively and these numbers rapidly decrease as more choice alternatives are considered, as the number of intransitive relations compared to transitive relations increases as a function of choice set size \cite[e.g.,][]{regenwetter2021ir}.} and enforcing such properties may lead to a highly restrictive and inflexible AI decision system.  Moreover, it remains an open question, whether such constraints truly benefit human users.  Recent work has challenged the notion that AI systems should be aligned with the preferences of a human user as described within utility frameworks, noting several shortcomings of applying rational choice theory to both humans and AI systems \citep{zhi2024beyond}.  Given these questions, there is a vast and still largely unexplored domain of AI rationality, underscoring the need for continued investigation. 

\appendix
\section{Appendix}
\subsection{Code Used for Experiment and Analysis}
All code for carrying out the experiment is available at \url{https://doi.org/10.6084/m9.figshare.28418687}

\subsection{Randomization Seeds Used for the Experiment} 
\begin{center}
    \begin{tabular}{|c|c|c|c|}
         835088831 &  420986496 &  698711259 &  208932753 \\
         622879400 &  647348309 &  675809175 &  819772621 \\
         675809175 & 429892269 \\
    \end{tabular}
    
\end{center}

\subsection{Question Examples}

\noindent\textbf{Monetary value + fraction}

Gamble 1 can give 25.43 with a chance of 7/24. Gamble 2 can give 24.16 with a chance of 1/3. Which do you choose?

\noindent\textbf{Monetary value + percentage} 

Gamble 1 can give 25.43 with a chance of 29.17\%. Gamble 2 can give 24.16 with a chance of 33.33\%. Which do you choose? 

\noindent\textbf{\$ Monetary value + fraction} 

Gamble 1 can give \$25.43 dollars with a chance of 7/24. Gamble 2 can give \$24.16 dollars with a chance of 1/3. Which do you choose? 

\noindent\textbf{\$ Monetary value + percentage}

Gamble 1 can give \$25.43 with a chance of 29.17\%. Gamble 2 can give \$24.16 with a chance of 33.33\%. Which do you choose? 

\noindent\textbf{Monetary value ‘dollars’ + fraction} 

Gamble 1 can give 25.43 dollars with a chance of 7/24. Gamble 2 can give 24.16 dollars with a chance of 1/3. Which do you choose? 

\noindent\textbf{Monetary value ‘dollars’ + percentage}

Gamble 1 can give 25.43 dollars with a chance of 29.17\%. Gamble 2 can give 24.16 dollars with a chance of 33.33\%. Which do you choose? 

\subsection{Prompt Examples}
\noindent\textbf{Instruct model example prompt}
\begin{lstlisting}[numbers=none, basicstyle=\small, breaklines=true]
<|begin_of_text|><|start_header_id|>system<|end_header_id|>
You have the choice of two gambles. Pick which one you would prefer.
<|eot_id|><|start_header_id|>user<|end_header_id|>
Gamble 1 can give 5.0 with a chance of 7/24. Gamble 2 can give 4.75 with a chance of 8/24. Which do you choose?<|eot_id|><|start_header_id|>assistant<|end_header_id|> I choose Gamble //
\end{lstlisting}    

\noindent\textbf{Base model example prompt}
\begin{lstlisting}[numbers=none, basicstyle=\small, breaklines=true]
You have the choice of two gambles. Pick which one you would prefer.
Gamble 1 can give 25.43 with a chance of 29.17%. Gamble 2 can give 24.16 with a chance of 33.33%.
Which do you choose?
I choose Gamble 
\end{lstlisting}

\subsection{Example of an AI Hallucination}
\noindent\textbf{Meta-Llama-3-8B}
\begin{lstlisting}[numbers=none, basicstyle=\small, breaklines=true]
You have the choice of two gambles. Pick which one you would prefer.
Gamble 1 can give $22.89 dollars with a chance of 9/24. Gamble 2 can give $19.32 dollars with a chance of 10/24.
Which do you choose?
I choose Gamble 1.
I choose Gamble 2.
I choose neither.
I don't know what to choose.
I can't choose.
I won't choose.
I choose Gamble 1 because I know that it will give me $22.89 dollars with a chance of 9/24. I choose Gamble 2 because I know that it will give me $19.32 dollars with a chance of 10/24.
I choose Gamble 1 because I know that it will give
\end{lstlisting}

\subsection{University-Based HPC Facility}
For the sake of computational power, this work was performed on the high-performance computing infrastructure provided by the Research Support Solutions and in part by the National Science Foundation under grant number CNS-1429294 at the University of Missouri, Columbia MO. DOI: https://doi.org/10.32469/10355/69802 .
\cleardoublepage

\end{document}